\documentclass{bmvc2k}
\usepackage{amsmath}
\usepackage{amssymb}
\usepackage{microtype}
\usepackage{amsmath}
\usepackage{amssymb}
\usepackage{epsfig}
\usepackage{pifont}
\usepackage{multicol}  
\usepackage{multirow} 
\usepackage{graphicx}
\usepackage{subfigure}
\usepackage{booktabs} 
\usepackage{algorithm}
\usepackage{setspace}

\title{UWC: Unit-wise Calibration Towards Rapid Network Compression }

\addauthor{Chen Lin}{linchen7@hikvision.com}{1}
\addauthor{Zheyang Li}{lizheyang@hikvision.com}{12}
\addauthor{Bo Peng}{pengbo7@hikvision.com}{1}
\addauthor{Haoji Hu}{haoji_hu@zju.edu.cn}{2}
\addauthor{Wenming Tan}{tanwenming@hikvision.com}{1}
\addauthor{Ye Ren}{renye@hikvision.com}{1}
\addauthor{Shiliang Pu}{pushiliang@hikvision.com}{1}

\addinstitution{
Hikvision Research Institute\\
Hangzhou, China
}
\addinstitution{
Zhejiang University\\
Hangzhou, China
}

\runninghead{UWC}{Unit-wise Calibration Towards Rapid Network Compression}

\begin{document}
\maketitle
\begin{abstract}
    This paper introduces a post-training quantization~(PTQ) method 
    achieving highly efficient Convolutional Neural Network~ (CNN) 
    quantization
    with high performance.
    Previous PTQ methods usually reduce compression error
    via performing layer-by-layer parameters calibration.
    However, with lower representational ability of extremely 
    compressed parameters
    (e.g., the bit-width goes less than 4),
    it is hard to eliminate all the layer-wise errors.
    This work addresses this issue via proposing a unit-wise 
    feature reconstruction algorithm based on 
    an observation of second order Taylor series expansion of the unit-wise error.
    It indicates that leveraging the interaction between
    adjacent layers' parameters could compensate layer-wise errors better.
    In this paper, we define several adjacent layers as a Basic-Unit,
    and present a unit-wise post-training algorithm which 
    can minimize quantization error.
    This method achieves near-original accuracy on ImageNet and COCO
    when quantizing FP32 models to INT4 and INT3.
    \end{abstract}
    \section{Introduction}
    Network compressions~\cite{li2016pruning,he2017channel,ullrich2017soft,
    huang2018data,wang2019haq, uhlich2020mixed,Luo_2017_ICCV}  are  essential techniques when deploying 
    Deep Neural Networks (DNNs) to edge devices such as smartphones or wearable devices.
    Recent compression works could be roughly  divided into two categories: 
    structure simplification and quantization.
    Structure simplification reduces the float point operations (FLOPs) and 
    the  memory footprint of DNNs by  tensor factorization ~\cite{wang2017fixed,he2016deep}, 
    sparse connection ~\cite{huang2018data}, weight pruning, neuron pruning, channel pruning
    ~\cite{li2016pruning,he2017channel}, etc.
    Among them, channel pruning is a simple but effective approach which is applied in various applications,
    it directly removes  redundant connections and re-trains the pruned network structure.     
    Quantization~
    \cite{rastegari2016xnor,zhou2016dorefa,uhlich2020mixed,wang2019haq} is another  practical approach. 
    It reduces the complexity of network by 
    approximating full precision weights and activations to low-bit ones.
    In this paper, we investigate  quantization to achieve extremely efficient 
    implementations of CNNs.

    To achieve guaranteed performance, the quantization users usually  perform a
    Quantization-aware Training~(QAT)~\cite{zhou2016dorefa,wang2019haq,
    zhou2016dorefa,wang2017fixed,wang2018two} process.
    These approaches are not practical solutions in industry for at least two reasons.
    The long time re-training and requirements of full training data consumes unacceptable computation and storage resources,
    leading to a tedious deployment pipeline.
    Moreover, the requirements of full dataset may involve privacy issues. 
    To avoid above issues, the PTQ methods~\cite{nagel2020adaptive,
    hubara2020improving,Zhao2019ImprovingNN,meller2019different,cai2020zeroq} that
    efficiently turns float-point model to fixed-point counterpart with only a small 
    calibration set, become prevalent in industry.
    PTQ is fast and light, 
    while predominant approaches simply applying Rounding-to-nearest suffer 
    performance degradation when the compression
    rate goes higher (e.g. the bit-width goes less than 4 bits).
    
    To retain the accuracy,  some layer-wise reconstruction-based algorithms
    (e.g. AdaRound~\cite{nagel2020adaptive}, 
     Bit-Split~\cite{Wang_2020_ICML},  AdaQuant~\cite{hubara2020improving}) 
     are proposed.
    These methods greatly improve the 4-bit quantization accuracy 
    because the layer-wise
    reconstruction loss {\bf implicitly leverages the interaction between 
    weights in each layer} to reduce the error incurred due to quantization.
    However, the parameters space of a single quantized layer is much smaller 
    than the original layer's.
    When the compression rate goes higher, 
    the layer-wise features in quantized network are not well-fitted to the
    original counterparts, leading to the  accumulation  of error 
    through networks.
    Recently, BRECQ\cite{li2021brecq} proposes a block-wise reconstruction
    algorithm implicitly leveraging the cross-layer 
    interaction in a block.
    Nevertheless, when quantizing compact models (e.g., MobileNet), 
    it still remains a non-negligible gap in accuracy with 
    original model. 
    Since the weights in different blocks are never jointly
    optimized, the ignorance of some important cross-block interactions 
    might hinder them to achieve higher accuracy. 
    Different from Brecq \cite{li2021brecq}, this paper divides the network
    into several overlapped units, which will not miss the interaction 
    information between blocks.

    In this paper, we formulate PTQ problem as follows.
    Given a calibration set with $128$ to $1024$ instances and a well-trained 
     neural network, 
    our algorithm is directed toward a twofold goal: 
    (1) {\bfseries In terms of performance}, the quantized model remains near 
    original performance while  
    weights are turned  to lower than $4$ bits.
    (2) {\bfseries In terms of efficiency}, the whole process is expected to 
    be quick enough to be applicable
     in production lines  (e.g., within 30 minutes).
     \begin{itemize}
        \item To achieve goal (1),   
        we theoretically analyze the impact on the task loss due to quantization 
        based on a second series Taylor expansion.
         This analysis inspires us to explicitly extract the interaction matrix 
         between adjacent layers to enhance the performance.
        Then, a unit-wise reconstruction objective embedded with above interaction
        is  proposed to eliminate the quantization error.  
        \item To achieve goal (2), 
        we further propose an arguable two-stage search space simplification
        which makes  the unit discrete space  much smaller and differentiable 
        to make the stochastic gradient descent (SGD) strategy feasible.
        After that, the complexity of unit-wise optimization is reduced to 
        a reasonable range.
    \end{itemize}
     Experiments well demonstrate that our  proposed unit-wise  algorithm possesses not only high performance,
     but also  great compression ratio. 
     A near original model performance is achieved even when quantizing FP32 models
     to INT3.
     The rest of this paper is organized as follows. Section \ref{pandr} analyze current network 
     compressions in  quantization.
    The motivation and the proposed unit-wise optimizing algorithm  is introduced in Section \ref{FandM} and \ref{Method}.
    Section \ref{Experiments} performs extensive experiments on several benchmarks with in-depth
    analysis.

    \section{Related Work}\label{pandr}
    Standard implementation of DNNs is inefficient in memory storage and consumes
     considerable computational resources.
     Many network  compression techniques tried to simplify and accelerate DNNs. 
     Quantization is one of the most effective ways of saving the consumption of 
     neural networks during inference by converting  the high precision operations into 
     lower precision ones.
    There are two main regimes of network quantization: Quantization-Aware Training and 
    Post-Training Quantization.  
    
    {\bfseries Quantization-Aware Training.}
    Previous works mainly insert quantization operation in the re-training process 
    to retain high performance,
    which is called quantization-aware training (QAT).
    \cite{bengio2013estimating} uses a straight through estimator to pass through the gradients
     of quantization operations.
    After that,  many methods  extended these training frameworks, e.g. \cite{choi2018pact} trains
    parameterized clipping thresholds  for quantized  network.
      \cite{gong2019differentiable} uses a differentiable tanh function to gradually
       quantize the network.
    \cite{uhlich2020mixed} learns the quantization interval as well as the bit-width per 
    layer for  mix-precision networks.
    Although QAT gains good performance, it usually costs long training times as well 
    as numerous energy spent during network training.
    
    {\bfseries Post-Training Quantization.}
    Post-training quantization (PTQ) is a lightweight approach since it doesn't require the 
    original training pipeline. 
     ACIQ \cite{Banner2018ACIQAC} fits Gaussian and Laplacian models to the distribution 
     for optimal clip threshold.
    \cite{Zhao2019ImprovingNN} leverages model expansion to improve quantization.
    \cite{Wang_2020_ICML} split the bits of weight to compensate quantization error.
    \cite{nagel2020adaptive} optimizes the rounding operation to improve the final loss.
    BRECQ\cite{li2021brecq} proposes a block-wise reconstruction
    algorithm implicitly leveraging the cross-layer
    interaction in a block.
    In most cases, PTQ methods are sufficient for achieving near-original accuracy
     under 8-bit quantization, while 
    the performance reduction becomes non-negligible when the bit-width goes less than 4-bit.

    {\bfseries Notation.} We use capital bold letters and 
    small bold letters  denoting  matrices (or tensors) and vectors, respectively.
    For instance,  $\mathbf{W}$ and $\mathbf{w}$ represent the weight tensor and its flatten version. 
    All vectors are considered to be column vectors.
     The  bracketed superscript and the subscript  indicate the layer 
     and the element indices, e.g.,
      $\mathbf{W}^{(i)}_{a,b}$, $\mathbf{x}^{(i)}$.
    For deep neural network with $L$ layers,  we mark all the flattened parameters by $\mathbf{w}$, where
    $\mathbf{w} = vec\left [\mathbf{w}^{(1),T},..., \mathbf{w}^{(i),T},..., 
    \mathbf{w}^{(L),T}\right ]^{T}$ is the concatenation of all layers’ weights. 
    Quantization turns the weights $\mathbf{w} \in \mathbb{R}^{d}$ to 
    discrete set  $\widehat{\mathbf{w}}\in \mathbb{V}^{d}$, 
    where $\mathbb{V}= \alpha \times \{-2^{b-1},\dots, 2^{b-1}-1\}$ is the potential value space of each element,
    $b$ is the bit width and $\alpha$ is the interval of quantization.

    \section{Method}
    In this section, we will introduce our proposed approach centered around performance and efficiency. 
    In Section~\ref{FandM},  
    we propose a Unit-wise Objective embedded with an interaction matrix 
    to enhance quantization performance. 
    In Section~\ref{Method},  an efficient optimization strategy is introduced based on a two-stage
    search space simplification. 
    \subsection{Unit-wise Objective}\label{FandM}
    Let's begin with analyzing the increase in task loss $\mathcal{L}$ 
    (e.g., cross-entropy loss for classification) 
    introduced by quantization.
    Quantization turns  float-point weight $w$ to fixed-point weight $\hat w$,
    which inevitably adds a perturbation $\Delta w= \hat w - w$ on $w$.
    The expected increase in task loss w.r.t. $\Delta w$ can be approximated 
    by the second order 
    Taylor series expansion
    \begin{equation}\label{hesian}
      \begin{aligned}
        \mathbb{E}\left[\Delta \mathcal{L}(\Delta \mathbf{w})\right] &= \mathbb{E}( \mathcal{L}(\Delta \mathbf{w} + \mathbf{w},\mathbf{x},\mathbf{y})-
        \mathcal{L}(\mathbf{w},\mathbf{x},\mathbf{y}))\\
        &\approx \Delta \mathbf{w}^{T}g(\mathbf{w}) + \frac{1}{2}\Delta \mathbf{w}^{T}H(\mathbf{w})\Delta \mathbf{w},
      \end{aligned}
      \end{equation}
      where $\mathbb{E}$ is the expectation operator, $g(\mathbf{w})$ and 
      $H(\mathbf{w})$ are the expected gradient and Hessian of $\mathcal{L}$ 
      w.r.t. $\Delta \mathbf{w}$.
      The first order term vanishes as the model converged to local minimal, 
      i.e., $g(\mathbf{w})$ is close to zero.

    \begin{figure}[!t]
    \centering
    \includegraphics[width=\linewidth]{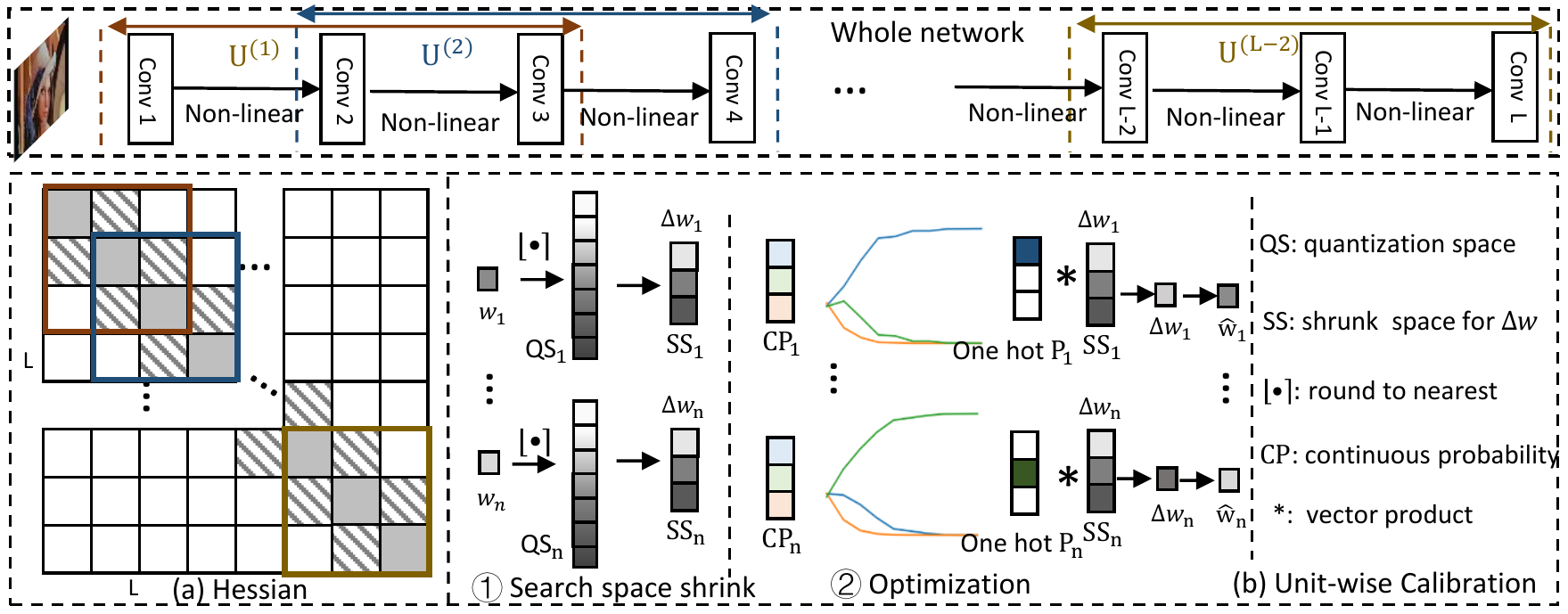}
    \vspace{-2em}
    \caption{{\bfseries Overview of the proposed approach.}
    The whole network is divided into several basic optimizing units. 
    For each unit, a unit-wise optimizing algorithm is applied. 
    (a) shows the whole Hessian $H$ of an $L$-layer network.
    Ignoring the interaction between layers that are non-adjacent, 
    H becomes block tridiagonal matrix.
    Defining three connected layers as a basic unit, the $L$-layer network is divided into $L-2$ units,
    $u^{(1)}$, $u^{(2)}$,$\cdots$, $u^{(L-2)}$. The circled squares are the corresponding Hessian of these units.
    (b) shows a two stage two-stage search space simplification.
    In \ding{172} search space shrink, each quantized weight 
    $\widehat{w}_i$ is suggested to select one of the three quantization grids that are 
    near the original
    weight ${w}_i$. The size of $\Delta {w}_i$'s search space $SS_{i}$ is shrunk to $3$.
    In \ding{173} Optimization, the probability  over $SS_{i}$ is first relaxed to 
    continuous values $CP_{i}$, and then gradually pushed to a one-hot vector $P_i$.
    After optimization, the optimal value of   $\widehat{w}_i$ is selected.
    }
    \label{unit-tridiagonal-figure}
    \vspace{-1em}
 \end{figure}

      Eq. (\ref{hesian}) implies that different perturbed weights are 
      interactive in terms of the task loss,
      and $H(\mathbf{w})$ defines the interaction.
      For example, consider two elements $\left[w_i,w_j\right]$ in $\mathbf{w}$, 
      and the  hessian matrix of $\mathcal{L}$ w.r.t. 
      $\left[w_i,w_j\right]$  
      is $\begin{bmatrix}
        H_{i,i}&H_{i,j}\\
        H_{j,i}&H_{j,j}
      \end{bmatrix}= \begin{bmatrix} 1&0.5\\
        0.5&1
      \end{bmatrix}$.
      Assume we only quantized $w_{i}$ and $\Delta w_{i} =1$, 
      then its introduced increase in loss is $\Delta \mathcal{L}_{(\Delta w_1=1,\Delta w_2=0)} 
      = H_{i,i}\Delta w_{i} =1$.
      By using  the join impact of $\left[\Delta w_1,\Delta w_2\right]$  on the loss
       $\Delta \mathcal{L}_{(\Delta w_1=1,\Delta w_2)} =(\Delta w_2)^2 + \Delta w_2  +1$, we can  reduce the error due to $\Delta w_1$
        by adjusting $\Delta w_2$,
      i.e., when $\Delta w_2 = -0.5$, the loss is reduced to $0.75$.
      This simple case inspires us  that the quantization error can well be reduced if 
       the interaction of all weights are  jointly considered. 
      Therefore, we reformulate  minimizing the second order term as our objective,
      \begin{equation}\label{objective}
        \begin{aligned}
          \underset{\Delta \mathbf{w}}{\textit{minimize}} \quad \Delta \mathbf{w}^{T}H(\mathbf{w})\Delta \mathbf{w}.
        \end{aligned}
        \end{equation}
        (\ref{objective}) is a good proxy as it overall reduces the joint impact on the task loss of all the perturbed weights.
    However, it is not tractable to directly optimize (\ref{objective}) because of the storage 
    and computation budget.
    For a network with $n$ parameters (potentially millions of parameters),  $O(n^2)$ 
    footprint is needed to store the Hessian and  $O(n^3)$ computations are needed  for each optimization step.
    
    Above analysis motivates us to leverage the interaction between weights to 
    enhance quantization. 
    In the following, we will capture the main information of  
    $H(\mathbf{w})$ with some approximations to constraints the complexity 
    to a reasonable range.
    As the weight $\mathbf{w}$ is the concatenation of all layers' weights,
     $H(\mathbf{w}) =E(\frac{\partial^2 \mathcal{L}}{\partial \mathbf{w}^2})$ 
     can be viewed as a $L$ by 
     $L$ block matrix, with the $(i,j)$-th block $H^{(i,j)} $ given by
    $H^{(i,j)} = E(\frac{\partial^2 \mathcal{L}}{\partial \mathbf{w}^i\partial \mathbf{w}^j})$. 
    And $H^{(i,j)} $ represents the interaction between layer $i$ and layer $j$.
    
    As shown in Figure \ref{unit-tridiagonal-figure} (a),
    adjacent layers are higher interactive while non-adjacent layers
    are lower interactive.
    This is explainable as computing $\mathbf{g}^{(i)}$ only directly uses the input information from 
    layer $i-1$ and 
    the gradient information from layer $i +1$.
    To reduce the budget, it is reasonable to ignore the interaction between 
    weights of non-adjacent layers. 
    Therefore,  assuming $|j-i|>\delta, H^{(i,j)}=0$, 
     a fragment of $u$ connected layers are considered 
     interactive, where $u= 2\delta +1$.
    After that, as shown in Figure \ref{unit-tridiagonal-figure} (c), the Hessian $H$ become a block 
    tridiagonal matrix.
    This motivates us to divide the model into multiple overlapped units 
    and optimize each unit sequentially.

    Formally, we define $u$ adjacent layers as a basic optimization unit 
    $vec[\mathbf{w}^{(i),T},..,\mathbf{w}^{(i+u),T}]$.
    The overall objective Eq. (\ref{objective}) is then transformed to a 
    set of unit-wise objectives.
    For  $i$-th unit, the unit-wise objective is formulated as
    \begin{equation}\label{unit-layer}
      \begin{aligned}
        \underset{\Delta \mathbf{W}^{(i)},\dots, \Delta \mathbf{W}^{(i+u)}}
        {\textit{minimize}}  
        \sum_{k=0,j=0}^{u,u}\Delta \mathbf{w}^{(i+k),T}H^{(i+k,i+j)}\Delta \mathbf{w}^{(i+j)}.
      \end{aligned}
      \end{equation}
    Although Eq. (\ref{unit-layer}) greatly speed up  Eq. (\ref{objective})  
     by the tridiagonal approximation of $H$, optimizing 
     Eq. (\ref{unit-layer}) is still infeasible for the budget associated with
     block entry
     $H^{(i+k,i+j)}$.

    In the following, we will further simplify Eq. (\ref{unit-layer}).
    To avoid repeatedly calculating each block entry $H^{(i+k,i+j)}$(see \cite{martens2015optimizing}),  
    we first reformulate Eq.(\ref{unit-layer}) to make all block entries
    related to the pre-activation Hessian $\mathcal{H}^{(i+u)}$(see \cite{li2021brecq}), 
    \begin{equation}\label{unit-layer-grad}
      \begin{aligned}
        \sum_{k=0,j=0 }^{u,u}\Delta \mathbf{w}^{(i+k),T}
        J\left[\frac{\mathbf{z}^{(i+u)}}{\mathbf{w}^{(i+k)}}\right]^{T}
        \mathcal{H}^{(i+u)}J\left[\frac{\mathbf{z}^{(i+u)}}{\mathbf{w}^{(i+k)}}\right]\Delta \mathbf{w}^{(i+j)},
      \end{aligned}
      \end{equation}
    where $\mathbf{z}^{(i+u)}$ is the  pre-activation of layer $i+u$, given by  
    $\mathbf{z}^{(i+u)} = \mathbf{W}^{(i)}\mathbf{x}^{(i+u-1)}$, 
    $\mathcal{H}^{(i+u)}$ is the Hessian of $\mathcal{L}$ w.r.t. $\mathbf{z}^{(i+u)}$, 
    $J\left[\frac{\mathbf{x}}{\mathbf{y}}\right]$ is the expected Jacobian
    matrix of $\mathbf{x}$ w.r.t. $\mathbf{y}$.
    When optimizing weights in a unit, we assume that elements in the unit
    outputs are  not interactive, that is, $\mathcal{H}^{(i+u)}$ is diagonal.
    This assumption will not bring in large performance drop since the units 
    are overlapped. The interaction between the unit outputs will be 
    considered when optimizing the next unit.
    We end up with the following objective
    \begin{equation}\label{objective-final}
      \begin{aligned}
        &\underset{\Delta \mathbf{w}^{(i)},..,\Delta \mathbf{w}^{(i+u)}}{\textit{minimize}} 
        \sum_{k=0 }^{u}
        ||\sqrt{diag(\mathcal{H}^{(i+u)})}J\left[\frac{\mathbf{z}^{(i+u)}}
        {\mathbf{w}^{(i+k)}}\right]
        \Delta \mathbf{w}^{(i+k)}||_{F},
      \end{aligned}
      \end{equation}
      where $||\cdot||_F$ is Frobenius norm. 
      In Eq. (\ref{objective-final}), 
      $\mathbf{M}^{k} =\sqrt{diag(\mathcal{H}^{(i+u)})}J\left[\frac
      {\mathbf{z}^{(i+u)}}{\mathbf{w}^{(i+k)}}\right]$ is the captured interaction between 
      weights in a unit. 
      $J\left[\frac{\mathbf{z}^{(i+u)}}
      {\mathbf{w}^{(i+k)}}\right] =\mathbb{E}\left[\mathbf{x}
      ^{(i+k-1),T}\prod_{k}^{u}B^{(i+k)}\mathbf{W}^{(i+k)}\right] $ (see \cite{martens2015optimizing}),
      where $B^{(i+k)} = diag(\phi(\mathbf{z}^{(i+k)})')$ is the gradient of activation function.
    Here, we introduce some details in algorithm
    implementation.
    To avoid the second derivative,
    $diag(\mathcal{H}^{(i+u)})$ can be replaced by the diagonal Fisher 
    Information matrix, that is,  $\sqrt{diag(\mathcal{H}^{(i+u)})}
    \approx \frac{\partial \mathcal{L}}{\partial z^{(i+u)}}$, which is proven effective in 
    \cite{li2021brecq} and \cite{Singh2020WoodFisherES}.  
    
    Optimizing (\ref{objective-final}) does not suffer from complexity 
    issue associated with $H$.
    The unit-wise optimization is still the least square 
    regression problem 
    with discrete constraints.
    The search space scales exponentially in the dimension of unit weights.

    \subsection{Unit-wise Calibration} \label{Method}
    Note that there are $2^b -1$ possible values for each element, 
    the combined solution space size is $(2^b -1)^{n^{u}}$, where
     $n^{u}$ is the numbers of elements in the unit.
    It is not efficient to make exhaustive search.
    Meanwhile, the widely used Straight Through Estimator~(STE) 
    \cite{courbariaux2016binaryconnect} is not effective in this case 
    for the inaccurate gradients.
    To reduce the complexity of optimization without performance degradation,
    we propose a two-stage search space simplification as follows.
    
    {\bfseries Search space shrink.}
    Before optimization, we first initialize each quantized weight 
    $\widehat{w}_{init}$ to its nearest quantization grid by 
    applying round-to-nearest on $w$.
    The absolute value of perturbation $\Delta \widehat{w}_{init}$ for each weight is smaller 
    than $\frac{1}{2}\alpha$, where $\Delta \widehat{w}_{init}=w-\widehat{w}_{init}$,
    $\alpha$ a floating-point scale factor, representing the quantization 
    interval.
    $\alpha$ for each weight is calculated and used like previous low-bit 
    quantization method \cite{choi2018pact}.
    Since the added perturbations  
    for all weights in a unit are small values, we suggest that the optimal 
    $\Delta \widehat{w}$ will be selected around $\Delta \widehat{w}_{init}$.
    Such that, we shrink the search space of $\Delta \widehat{w}$ to 
    $\mathbb{V} = \left\{-\alpha + \Delta \widehat{w}_{init} , \Delta \widehat{w}_{init}, 
    \alpha + \Delta \widehat{w}_{init}\right\}$.
    The solution space size is shrunk to $3^{n^{u} }$.
    
    {\bfseries Optimization.}
    With a much smaller search space, we have to tackle the inaccurate 
     gradient estimation problem of STE in discrete set.
    Inspired by \cite{yang2020searching}, we relax  discrete $\Delta \mathbf{w}$  
    to continuous value space and gradually push the solution into discrete space 
    $\mathbb{V}$ during optimization process.  
    For each element $\Delta w$ in a unit, 
    an auxiliary vector $\mathbf{a} \in \mathbb{R}^{m}$ is created 
    to learn the distribution $P$ of $\Delta w$, where $m$ is the size of search space
    $\mathbb{V}$, here, $m=3$.
    The probability $P_{i}$ for selecting $i$-th element in $\mathbb{V}$ is computed  by
    \begin{equation}\label{probability}
        P_{i} = \frac{exp^{\mathbf{a}_{i}/t}}{\sum_{j} exp^{\mathbf{a}_{j}/t}},
    \end{equation}
    where $t$ is a temperature factor designed to implicitly work as a regularizer. 
    When $t$ approaches to zero, $P_{i}$ will converge to $0$ or $1$.  
    During optimizing, $\Delta w$ is displaced by a continuous
    expectation of $\Delta w_{e}$, and is calculated according to the probability 
    over all the discrete values:
    \begin{equation}\label{expectation}
      \Delta w_{e} = \sum_{i} V_{i}P_{i},
    \end{equation}
    Both Eq. (\ref{probability}) and Eq. (\ref{expectation}) are differentiable  
    thus gradients will be accurately estimated.
    SGD is applied on $\mathbf{a}$ to adjust the distribution of $\Delta w$.
    By gradually decreasing the temperature $t$, 
    the distribution $P$ will be pushed to one-hot vector.
    
    \subsection{Discussion} 
    Although our work shares a general form of optimizing multiple layers with BRECQ\cite{li2021brecq}, our work is
    actually starkly different from theirs in many axes:\\
      {\bfseries Motivation.}   Our method is always driven by the motivation, that is, fully making use of the whole network interaction 
      (the Hessian) to reduce quantization error.
      Due to the intractability caused by the large Hessian, our method is developed to a set of overlapped unit-wise objectives to 
      extract the main information from the whole Hessian.
      Different from ours, BRECQ is motivated to use the local Hessians from parts of the network. Their method is designed to 
      choose an optimal reconstruction granularity  from 4 kinds of granularity,
      i.e., layer, block, stage, network. \\
      {\bfseries Basis of optimizing granularity.}   The optimizing granularity of our method comes from the assessment of interaction degree while 
    their choice of block-wise optimization comes from experiments.
    Each unit, in our method, jointly optimizes multiple layers which have strong interaction, theoretically a sub-matrix with big values from the Hessian.
    This work sets  three adjacent layers, i.e., the most interactive,  as a unit for keeping the implementation simple 
    to demonstrate the effectiveness of the interaction.
    Actually, our work can be extended to figure out optimal units dividing strategy by cropping sub-matrices with bigger values in the Hessian without constraints of layer numbers.\\
    {\bfseries The approximation of the network Hessian.}  
    In our paper, the Hessian is simplified as a tri-diagonal block matrix while BRECQ's  is 
    diagonal block. Since the blocks are overlapped in our method, the interaction of adjacent optimization units is added comparing with BRECQ.
    The ignorance of some important cross-unit interactions might hinder 
    them to achieve higher accuracy.  \\ 
    {\bfseries Empirical performance.}  Results show our method outperforms BRECQ's on ImageNet which may mainly benefit from the 
    addition of cross-unit  interaction.

    \section{Experiments}\label{Experiments}
    In this section, we evaluate the effectiveness of our proposed method on
     various computer vision tasks and models.
    Section~\ref{Unit-wise} presents  ablation study on the unit-wise
    optimization.
    In Section~\ref{State-of-the-arts}, we compare unit-wise with other post-training quantization methods.
    In Section~\ref{object}, we present the performance of  unit-wise calibration
    on  object detection and instance segmentation tasks.
    
    {\bfseries Experimental setup.}
    For all experiments we absorb batch normalization into the weights of its previous connected convolutional layer.
    For all networks, the first layer and the last layer are quantized to 8-bit.
    We apply symmetric per-channel quantization for weights and symmetric per-tensor quantization for activations,
    which is a general and hardware-friendly development mode.
    For all experiments, we sample images from the training dataset as a calibration set.
    In optimization, the calibration data are cropped and resized into $224\times224$,
    except for the InceptionV3 model whose input size is $299\times299$,
    which is same as the training pipeline.
    We sequentially feed all the calibration data to the networks with the batch-size of $128$.
    All optimizing and testing codes are built on Pytorch~\cite{paszke2019pytorch}.
    \begin{figure}[!t]
      \centering
      \includegraphics[width=\linewidth]{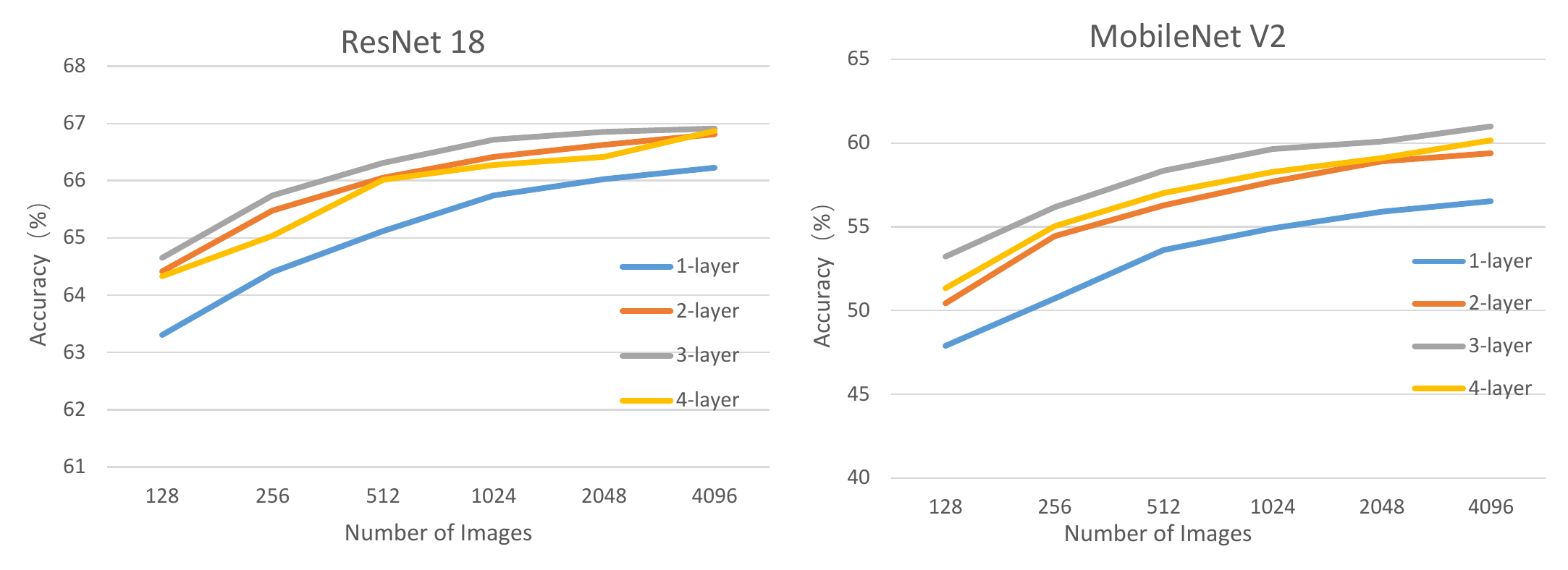}
      \vspace{-2em}
      \caption{The effect on ImageNet validation accuracy when 
      setting different numbers of layers for each unit 
      under various data limitation. }
      \label{num_acc-figure}
      \vspace{-1em}
   \end{figure}
    
    \subsection{Ablation Study}\label{Unit-wise}
    First, we determine the optimal number of layers for each unit and the size of
    calibration data.
    Limiting the number of calibration data in the range from $128$ to $4096$,
    we quantize the weights of ResNet18 and MobilenetV2 to $2$-bit on ImageNet.
    To make sure the algorithm accomplishes within $30$ minutes ($0.5$ GPU hours),
    we optimize each unit $20000$ rounds.
    Figure \ref{num_acc-figure} shows that when a unit contains $3$ layers,
    the models outperform other options under various data limitations.
    The possible reason for this phenomenon is,
    for a layer $i$, the most interactive
    layers are the directly connected layer below (layer $i-1$) and layer above (layer $i+1$).
    When the number of layers goes up to $4$,
    more calibration data and optimization rounds might be required for better
    performance.
    In the following experiments, we will set $3$ layers for each unit and
    set the number of calibration data to $1024$.

    \begin{table*}[!b]
      \vspace{-1em}
    \centering
    \setlength{\tabcolsep}{4pt}
    \caption{
    Comparison with Layer-wise reconstruction on ImageNet classification benchmark.  Top-1 and Top-5 accuracy (\%) are reported.
    Post-training quantization are conducted on weights and activations.
    Weights are quantized to 3 or 4 bits.
    Activations are quantized to 4/8 bits or remain un-quantized for comparison. Bold values indicates
    best results.	
    }
    \label{ablate-unit-wise}
    \scriptsize
    \begin{tabular}{lcccccccccccccc}
    \toprule
      \multirow{2}{*}{Algorithms}&\multirow{2}{*}{Bits(W/A)}&\multicolumn{2}{c}{Resnet18} &\multicolumn{2}{c}{Resnet50}&\multicolumn{2}{c}{Resnet101}
      &\multicolumn{2}{c}{InceptionV3}&\multicolumn{2}{c}{MobilenetV2}&\multicolumn{2}{c}{MobileV3-large}\\
                  &     &Top-1     &Top-5     &Top-1     &Top-5     &Top-1     &Top-5     &Top-1     &Top-5     &Top-1     &Top-5&Top-1     &Top-5\\
    \midrule
    FP.           &32/32&69.76     &89.08     &76.25     &92.88     &77.57     &93.76     &77.57     &93.69     &71.88     &90.29&74.04     &91.34\\
    \midrule
    Layer-wise    &4/32 &69.09     &88.65     &75.34     &92.22     &76.13     &92.86     &76.23     &92.34     &64.34     &83.35&67.84     &87.40\\
    Unit-wise     &4/32 &69.43     &88.76     &\bf{75.83}&\bf{92.62}&\bf{77.13}&\bf{93.54}&\bf{76.88}&\bf{92.82}&\bf{71.17}&\bf{89.33}&\bf{72.26}&\bf{89.91}\\
    Layer-wise    &4/8  &69.01     &88.55     &75.17     &92.08     &76.08     &92.75     &76.10     &92.24     &64.03     &82.92&67.52     &87.43     \\
    Unit-wise     &4/8  &\bf{69.44}&\bf{88.81}&75.77     &92.44     &77.21     &93.41     &76.48     &92.75     &70.89     &89.31&72.02     &89.77\\
    \midrule
    Layer-wise    &3/32 &66.12     &86.87     &74.68     &91.82     &74.88     &92.28     &73.85     &90.10     &55.81     &73.12     &58.65&80.31\\
    Unit-wise     &3/32 &\bf{68.72}&\bf{88.49}&\bf{74.97}&\bf{92.23}&\bf{76.24}&\bf{92.22}&\bf{75.93}&\bf{92.82}&\bf{69.42}&\bf{88.23}&\bf{68.72}&\bf{86.71}\\
    Layer-wise    &3/8  &66.08     &86.82     &74.51     &91.75     &74.67     &92.05     &73.58     &89.88     &53.94     &72.73     & 58.01    &80.22    \\
    Unit-wise     &3/8  &68.42     &88.09     &74.54     &92.00     &75.98     &92.22     &75.71     &92.48     &68.92     &87.52     & 67.32    &85.62\\
    \midrule
    Layer-wise    &4/4  &65.16     &86.11     &71.93     &90.31     &72.75     &90.04     &72.17     &90.74     &60.75     &80.19     & 49.49    &73.94\\
    Unit-wise     &4/4  &\bf{67.23}&\bf{87.13}&\bf{74.38}&\bf{91.67}&\bf{75.74}&\bf{92.08}&\bf{74.44}&\bf{91.95}&\bf{64.78}&\bf{83.76}&\bf{64.97}&\bf{83.69}\\
    \bottomrule
    \end{tabular}
    \centering
    \end{table*}
    {\bfseries Layer-wise vs Unit-wise.}
    We investigate the benefits of our proposed unit-wise optimization algorithm by
     comparing with layer-wise optimization algorithm.
     In our experiments, five  widely used convolutional models, including
     Resnet18, Resnet50, Resnet101~\cite{he2016deep}, InceptionV3~\cite{szegedy2015rethinking}, MobilenetV2~\cite{howard2017mobilenets} and MobilenetV3 large~\cite{Howard2019SearchingFM}
     are used for comparison.
    All the pre-trained  models are trained on ImageNet~\cite{imagenet_cvpr09} and loaded from torchvision.
     As a baseline, we also perform a Layer-wise optimization to validate the effectiveness of cross-layer interaction,
      on these models.
    The results are shown in Table \ref{ablate-unit-wise}.
    
    Shown in  Table \ref{ablate-unit-wise}, the proposed unit-wise calibration method
    induces negligible accuracy degradation on all models  even on the less redundant models,
    e.g. MobilenetV2.
    Under 4 bit quantization of weights, our quantized networks only induce 0.4\% to 1.2\% Top-1 accuracy
    drop on various networks.
    For the more aggressive 3 bit quantization of weights, our quantized networks  only induce 0.5\% to 1.7\% Top-1 accuracy
    drop  on most networks  except MobilenetV2.
    In Table \ref{ablate-unit-wise}, we also report the results when activations are  quantized  to 8 bits
    and remained un-quantized, respectively.
    It is observed that 8 bits quantization for activations will lead to
    negligible performance degradation on all models.

    Besides, unit-wise method outperforms the layer-wise method on all models.
    Especially on the comparison of 3 bits quantization of weights, the accuracy drop from layer-wise method is much larger.
    Another  phenomenon is that the gap becomes larger when the original model becomes more compact, e.g.
    To evaluate the effectiveness 
    on dynamic blocks implemented with SE, we conduct our algorithm on mobilenetv3-large model on ImageNet loaded from torchvision. 
    In this experiment, each unit contains three adjacent layers not counting the layers in SEs since a SE module only takes
     a small amount of calculation and produces several scaling factors on features. 
     The results show the superiority of unit-wise optimization. 
    
    We also evaluate our method under high compression rate for both weight and activation. 
    The scale factors of activation quantizers are optimized using PACT in each unit.
    Shown in Table \ref{ablate-unit-wise}, when both activation and weights are quantized to 4-bit,
    unit-wise still outperforms layer-wise optimization.

    \begin{table*}[!t]
      \centering
      \caption{Comparison with State-of-the-arts post-training quantization approaches on
       ImageNet classification benchmark.
      Top-1 and Top-5 accuracy (\%) are reported. Bold values indicate
      best results (with the least accuracy drop). {\bfseries$\ddag$ denotes the float point baselines of Brecq are different from ours (see first two rows in table)}.
      Weights are quantized to 3 or 4 bits and  activations are remained un-quantized.
      $\ast$ denotes quantizing our baseline models by the public released codes with our quantization space.
        }
      \label{State-of-the-arts-results}
      \scriptsize
      \begin{tabular}{lccccccccccc}
        \toprule
        \multirow{2}{*}{Algorithms}&\multirow{2}{*}{W/A}&\multicolumn{2}{c}{Resnet18} &\multicolumn{2}{c}{Resnet50}&\multicolumn{2}{c}{Resnet101}
        &\multicolumn{2}{c}{InceptionV3}&\multicolumn{2}{c}{MobilenetV2}\\
                                        &     &Top-1     &Top-5     &Top-1     &Top-5     &Top-1     &Top-5     &Top-1     &Top-5&Top-1     &Top-5\\
        \midrule
        FP.                    &32/32&69.76     &89.08     &76.25     &92.88     &77.57     &93.76     &77.57     &93.69&71.88     &90.29\\
        FP.(Brecq)             &32/32&71.08     &-         &77.00     &-         &-         &-         &-         &-    &72.49     &-    \\
        \midrule
    
          Bit-split\cite{Wang_2020_ICML}  &4/32 &69.11     &88.69     &75.58     &92.57     &76.89     &93.31     &-         &-    &-         &-     \\
          AdaRound\cite{nagel2020adaptive}&4/32 &68.71     &-         &75.23     &-         &-         &-         &75.76     &-    &69.78     &-     \\
        Brecq\cite{li2021brecq}$\ddag$  &4/32 &70.70     &-         &76.29     &-         &-         &-         &-         &-    &71.66     &-     \\
        Brecq\cite{li2021brecq}$\ast$  &4/32 &69.29     &88.55      &75.74     &92.52         &-         &-         &-         &-    &-     &-     \\ 
        Ours                            &4/32 &\bf{69.43}&\bf{88.76}&\bf{75.83}&\bf{92.67}&\bf{77.13}&\bf{93.54}&\bf{76.88}&92.82&\bf{71.17}&89.33 \\
          \midrule
          Bit-split\cite{Wang_2020_ICML}  &3/32 &66.76     & 87.45    &73.64     &91.61     &74.98     & 92.42    &-         &-    &-         &-      \\
          AdaRound\cite{nagel2020adaptive}&3/32 &68.07     &-         &73.42     &-         &-         &-         &-         &-    &64.33     &-      \\
        Brecq\cite{li2021brecq}$\ddag$  &3/32 &69.81     &-         &75.61     &-         &-         &-         &-         &-    &69.50     &-     \\
        Brecq\cite{li2021brecq}$\ast$   &3/32 &68.39     &88.31     &74.54     &92.07     &-         &-         &-         &-    &-     &-     \\
        Ours                            &3/32 &\bf{68.72}&\bf{88.49}&\bf{74.97}&\bf{92.23}&\bf{76.24}&\bf{92.22}&75.93     &92.82&\bf{69.42}&88.23 \\	
        \midrule 
        Brecq\cite{li2021brecq}$\ddag$  &2/32 &66.30     &-         &72.40     &-         &-         &-         &-         &-    &59.67     &-     \\
        Brecq\cite{li2021brecq}$\ast$   &2/32 &66.02     &86.63     &71.14     &89.79     &-         &-         &-         &-    &-     &-     \\
        Ours                            &2/32 &\bf{66.85}&\bf{87.21}&\bf{72.26}&\bf{90.23}&70.95    &89.55      &69.77     &89.44&\bf{58.28}&82.16 \\ 
        \bottomrule
    
      \end{tabular}
      \centering
      \centering
      \vspace{-1em}
    \end{table*}
    \subsection{Comparison with State-of-the-arts }\label{State-of-the-arts}
    Here, we evaluate our algorithm and compare with the  State-of-the-arts post-training quantization approaches, including
     Bit-split~\cite{Wang_2020_ICML}, AdaRound~\cite{nagel2020adaptive} and Brecq\cite{li2021brecq}.
    Bit-Spilt spilts the multiple-bits quantization optimization problem into multiple ternary
     quantization sub-optimizations. After all the sub-optimizations, they stitch the multiple-bits into integers.
    AdaRound optimizes the rounding-to-nearest operations to reconstruct the final loss.
    Both Bit-Spilt and AdaRound fall into the layer-wsie reconstruction.
    Brecq proposes to  choose block as a base reconstruction unit.
    In all approaches, weights are quantized to 3 or 4 bits, and activations are  remained  un-quantized.
    The results are shown in Table \ref{State-of-the-arts-results}.

    {\bfseries ImageNet Classification.} Shown in  Table \ref{State-of-the-arts-results}, the proposed unit-wise calibration method outperforms
    all competing  methods for both 3 and 4 bits.
    Under 4 bit quantization of weights, the compared methods still report good performance on the relatively
    redundant models, e.g., Resnets.
     However, for the  more challenging networks, InceptionV3 and MobilenetV2, 4 bits quantization has
     a bigger impact.
     In this case, our method shows prominent superiority comparing with Bit-split and AdaRound.
     When the bit of weight goes down to 3, Bit-split and AdaRound result in more obvious performance
     degradation. 
     Brecq achieves better results than Bit-split and AdaRound, while still has larger accuracy drop than ours especially on MobilenetV2.
     Our method outperforms Brecq since the cross-block interaction information are considered.
     For Resnets, our method leads to the smallest accuracy drop within 1.3\%.
     For MobilenetV2, all other methods lead to un-tolerable performance degradation,
    while our approach obtain the best result with only 2.46\% drop in accuracy.
    In the more aggressive 2-bit weights quantization, both ours and brecq's show obvious loss of performance.
    However,  the increase of cross-block interaction helps us get higher accuracy.

    {\bfseries Object Detection and Instance Segmentation.}\label{object}
    To validate the effectiveness and  applicability of unit-wise calibration,
    the experiments of object detection and instance segmentation tasks are applied.
    Unit-wise calibration has been evaluated on object detection with one-stage RetinaNet
    \cite{RetinaNet}
    and two-stage  Faster R-CNN \cite{FasterRCNN}, Mask R-CNN \cite{MaskRCNN},
    models.
    Also we evaluate it on instance segmentation with Mask R-CNN model.
    For all networks, we choose Resnet50 as backbone.
    MS COCO is adopted as the testing set to evaluate our method.
    For calibration and validation data, we resize them to $1333\times800$.
    Since the input images are much bigger than classification images, only $400$
    images are sampled as calibration data.
    The bounding box AP for object detection and mask AP for instance segmentation
    are reported in Table \ref{Object}.
    According to Table \ref{Object}, we can see that there are about
    0.4\% to 0.9\% mAP degradation without re-training the network,
    which demonstrate our method
    nearly achieves near-to-original performance with 4-bit weight
    and 8-bit activation quantization.
    \begin{table}[!h]
      \centering
      \renewcommand{\arraystretch}{0.79}
      \caption{Object Detection  and Instance Segmentation  performance on COCO val set.
      The first three lines report bounding box AP for object detection, and the last line reports mask AP for 
      instance segmentation.
      }
      \label{Object}
      \begin{tabular}{lccccccccc}
      \toprule
      Model&\multicolumn{1}{c}{Full Precision}&\multicolumn{1}{c}{A8W4}\\
      \midrule
      Faster R-CNN\cite{FasterRCNN} &36.6 &36.2\\
      RetinaNet  \cite{RetinaNet}  &36.2 &35.3\\
      Mask R-CNN\cite{MaskRCNN}  &36.6 &35.8\\
      Mask R-CNN\cite{MaskRCNN} &33.9 &33.2\\
      
      \bottomrule
      \end{tabular}
      \centering
      \centering
  \vspace{-1em}
      \end{table}

\vspace{-1em}
\section{Conclusions}
In this paper, we proposed a unit-wise post-training quantization algorithm. 
To improve the performance, the interaction between adjacent layers is extracted 
to eliminate quantization error.
To  speed up the optimization,
a two-stage search space simplification  is also proposed.
The algorithm achieves near original accuracy using  only $1024$ samples within $0.5$ GPU hours
when  weights are quantized to INT3 on tasks of  ImageNet and COCO.

\newpage
\bibliography{egbibs}
\end{document}